\documentclass[11pt,a4paper]{article}
\usepackage[hyperref]{emnlp-ijcnlp-2019}
\usepackage{times}
\usepackage{latexsym}
\usepackage{amsmath}
\usepackage{amssymb}
\usepackage{caption}
\usepackage{graphicx}

\usepackage{url}

\aclfinalcopy %

\definecolor{ramiblue}{HTML}{007bd5}
\definecolor{ramigray}{HTML}{a9a9a9}
\definecolor{ramired}{HTML}{c11b24}

\newcommand{\lmoneb}{\texttt{lm1b}}

\newcommand{\comment}[1]{}

\title{Bridging the Gap for Tokenizer-Free Language Models}

\author{Dokook Choe\thanks{~~Equal contribution.}\\
 \And
 Rami Al-Rfou\footnotemark[1]
 \And 
 Mandy Guo
 \\\hspace{0.5pt}
 \\
 Google \textsc{AI} \\
 1600 Amphitheatre Parkway\\
 Mountain View, California 94043 \\
 \{\tt choed, rmyeid, xyguo, hylee, nconstant\}@google.com
 \And
 Heeyoung Lee
 \And
 Noah Constant
}

\date{}

\begin{document}
\maketitle
\begin{abstract}
Purely character-based language models (LMs) have been lagging in quality on large scale datasets, and current state-of-the-art LMs rely on word tokenization.
It has been assumed that injecting the prior knowledge of a tokenizer into the model is essential to achieving competitive results.
In this paper, we show that contrary to this conventional wisdom, tokenizer-free LMs with sufficient capacity can achieve competitive performance on a large scale dataset.
We train a vanilla transformer network with 40 self-attention layers on the One Billion Word (\lmoneb{}) benchmark and achieve a new state of the art for tokenizer-free LMs, pushing these models to be on par with their word-based counterparts.
\end{abstract}

\section{Introduction}
There has been a recent surge of improvements in language modeling, powered by the introduction of the transformer architecture~\cite{transformer}.
These gains stem from the ability of the transformer self-attention mechanism to better model long context (as compared to RNN networks), spanning hundreds of characters~\cite{bytelmaaai} or words~\cite{baevski2018adaptive,radford2019language}.
These approaches consider language modeling as a classification problem with the aim of predicting the next token given a fixed-size preceding context.
To support variable-length context, \citet{transformerxl} adds recurrence to a transformer model, improving the state-of-the-art further.

Current word-based language models (LMs) depend on a series of preprocessing steps that include lowercasing, tokenization, normalization, and out-of-vocabulary handling.
This preprocessing stage is language dependent and can add significant complexity to real applications.
As such, it is appealing to shift to more general LMs that process raw text at the character level.
Processing language at the character level allows us to model morphological variants of a word, assign reasonable likelihood to out-of-vocabulary words, and learn subword-level language abstractions.
This open vocabulary modeling is quite important for languages with complex morphology such as Arabic, Turkish, or Finnish \cite{DBLP:journals/corr/LeeCH16, bojanowski-etal-2017-enriching, Kim:2016:CNL:3016100.3016285}.

While character- and word-based LMs have both improved in their performance over time, purely character-based LMs have continued to lag in performance compared to models that leverage a tokenizer.
\citet{bytelmaaai} report inferior performance from character-level modeling on a large scale word-level benchmark, \lmoneb{}~\cite{lm1b}.
Similarly, \citet{radford2019language}
observe that a character-level LM is harder to train to competitive performance on their huge WebText corpus, as compared with subword segmentation using byte pair encoding (BPE) \cite{sennrich-etal-2016-neural, gage1994compression}.

Sub-word tokenization approaches like BPE represent a middle ground for text segmentation.
On one hand, they can help with better modeling open vocabulary.
On the other hand, they still depend on a tokenizer, adding complexity to the final system.
Moreover, the preprocessing stage is not jointly optimized with learning the task objective.
This last point is especially relevant given that LMs are increasingly used for their ability to produce pretrained representations that will be fine-tuned for a downstream task \cite{bert,ulmfit,gpt,elmo}.
Since word-based LMs use closed vocabulary and sub-word models adopt a segmentation that targets the pretraining corpus, there is little space to adapt the vocabulary or optimize the segmentation to fit the final task data distribution.

The rest of this paper is organized as follows.
In Section \ref{sec:modeling}, we describe our model architecture, which is a vanilla deep transformer byte-level LM.
Section \ref{sec:dataset} describes the \lmoneb{} dataset and our evaluation methodology.
Section \ref{sec:results} presents our results and how our model compares to the previous work.
In Section \ref{sec:representations} we analyze the representations learned by the network at different depths using word-similarity benchmarks.
For this analysis to be feasible we propose a  strategy to extract word representations from a character model.

To summarize our contributions:
\begin{itemize}
    \item We develop a competitive tokenizer-free language model on a large scalable dataset.
    \item We probe the performance of our model's learned intermediate representations on word similarity tasks.
\end{itemize}

\section{Modeling}
\label{sec:modeling}

Language models (LMs) assign a probability distribution over a sequence $x_{0:t}$ by factoring out the joint probability from left to right as follows
\begin{equation}
    P(x_{0:t}) = P(x_0) \prod_{i=1}^{t} P(x_{i}|x_{< i}).
\end{equation}
Instead of reading in the tokenized input text, our model reads raw \mbox{utf-8} bytes.
For English text in the ASCII range, this is equivalent to processing characters as individual tokens.
Non-ASCII characters (e.g.~accented characters, or non-Latin scripts) are typically two or three \mbox{utf-8} bytes.
We use a standard ``transformer decoder'' (a stack of transformer layers with a causal attention mask) to process the sequence $x_{0:i-1}$ and predict the following byte $x_i$.
The model's prediction is an estimate of the probability distribution over all possible 256 byte values.
Our input byte embedding matrix has dimensionality 256.
Our byte-level transformer model has 40 standard transformer layers with hidden size 1024, filter size 8192, and 16 heads.
The model has around 836M parameters, of which only 66K are byte embeddings. 

\subsection{Training}
We sample random byte sequences of length 512.
This sampling process does not respect the sentence boundary.
Therefore, one example might span complete and partial sentences.
We dropout both timesteps of self-attention layers and features of relu activations across timesteps with a probability of 0.3.
We use the Adam optimizer~\cite{adam} with initial learning rate $10^{-4}$ and batch size 1024.
The training runs for two million steps, and at every 10,000 steps we decay the learning rate geometrically by 0.99.

\subsection{Windowed Prediction}
To score each byte prediction, we need to process an entire 512-byte context from scratch, which is computationally intensive. To speed up development, for each window of context size $c$, we score $(\text{stride}=c/2)$ characters in parallel (the second half of the window).
This leads to a tractable running time for our development evaluation process.
While this setup is sub-optimal for our model, we did not observe any significant regression in our metrics.
For example, the final bits/byte value of 0.874055 ($\text{stride}=1$) only grows to 0.87413 with $\text{stride}=256$.
Our final test evaluation is reported with $\text{stride} = 1$.

\begin{table}[b]
\begin{center}
\begin{tabular}{c|r r r}
    & shards &  \# of words & \# of bytes\\
    \hline
    train & 01-99 & 798,947,561 & 4,132,763,660 \\
    dev  &    01 of 00 & 165,560 & 856,742 \\
    test &    00 of 00 & 159,658 & 826,189 \\
\end{tabular}
\end{center}
\caption{Word and byte counts in \lmoneb{}.
The word count includes end of sentence tokens, and byte count includes newline characters.
}
\label{corpus}
\end{table}

\begin{table*}[!ht]
\begin{center}
\begin{tabular}{ l|c c c c c} 
 & Segmentation & Context Length & \# of params & Perplexity & Bits/Byte\\
 \hline
 \newcite{shazeer2017outrageously} & Word & Fixed & 6.0B & 28.0 & 0.929 \\
 \newcite{shazeer2018mesh} & Word-Piece & Fixed & 4.9B & 24.0 & 0.886 \\ 
 \newcite{baevski2018adaptive} &  Word & Fixed & 1.0B & 23.0 & 0.874 \\ 
 \newcite{transformerxl} & Word & Arbitrary
 & 0.8B & 21.8 & 0.859 \\ 
 \hline
 \newcite{bytelmaaai} & Byte & Fixed & 0.2B & 40.6 & 1.033 \\
 Ours & Byte & Fixed & 0.8B & 23.0 & 0.874 \\ 
\end{tabular}
\end{center}
\caption{Comparing recent language model results on \lmoneb.}

\label{result_table}
\end{table*}

\section{Experimental Setup}
\label{sec:dataset}

There are no large scale datasets that are heavily studied for both word and character language modeling.
Typically, a specific dataset will be considered under just one level of segmentation.
For our efforts to be comparable with the literature, we use a word LM dataset.
This puts our model at a disadvantage; the dataset is tokenized and our model will not utilize the given word boundary information.
Our approach is able to model rare words and estimate their appropriate likelihoods, however, they have been replaced with a special token to produce closed vocabulary text that is appropriate for word-level modeling.
Hence, the metrics we report are meant to provide a lower bound on the utility of our approach in realistic settings.

\subsection{LM1B}
We use the One Billion Word benchmark~\cite{lm1b} to compare LM performance.
The dataset consists of shuffled short sentences, and doesn't require modeling long contexts (95\% of the training sentences are under 256 bytes and over 99.78\% are under 512 bytes).
The corpus is tokenized, and a small percentage of rare words are replaced with UNK tokens.
The data gets split into 100 shards, and the first one (00) is held out while the rest (01-99) are used for training.
The holdout set is split again into 50 shards, and historically shard 00 of the holdout has been used as the test set.
There is no standard dev set, so we use shard 01 of the holdout as dev.
See the corpus statistics in Table~\ref{corpus} for details.

\subsection{Metrics}
Word LMs typically report their results in terms of perplexity per word $(ppl)$ while byte LMs report their results in bits per byte $(bpb)$.
We report both metrics to make our results more accessible.

Conversion between those metrics are based on the following observation: \emph{The amount of information in the test dataset is the same independent of segmentation.}

\begin{align}
  \label{eq:conv1}
  I(text) & = |words| \times \frac{bits}{word}  = |bytes| \times \frac{bits}{byte},
\end{align}
 where $I(x)$ is the information contained in $x$, which is $- \log_2 P(x; model)$.
Equation~\ref{eq:conv1} allows us to convert bits/word to bits/byte.
Then straightforwardly, using Equation~\ref{eq:ppl} we can convert $bpb$ to $ppl$:
\begin{equation}
\label{eq:ppl}
      perplexity = 2^{\frac{|bytes|}{|words|} \times bpb}
\end{equation}

We train our model to minimize $bpb$ over the training set and convert $bpb$ on the test set to $ppl$ for comparison.
For the test dataset, we use the $|words|$ and $|bytes|$ values reported in Table~\ref{corpus}.

\section{Results and Discussion}
\label{sec:results}
Table \ref{result_table} shows the perplexity of several models on \lmoneb{}.
We observe that tokenizer-free LM performance improves significantly (40.6 to 23.0) when the model capacity is increased from 0.2B to 0.8B parameters.
With sufficient capacity our byte-level LM is competitive with word based models (ranging from 21.8 to 28.0).
Note, our model is able to achieve comparable performance without any explicit signal of word boundaries.

Because of the large symbol space that word-based LMs address, they rely on sparse operations running on heterogeneous devices to run efficiently (e.g.~running sparse embedding lookups on CPU as opposed to GPU/TPU).
By contrast, byte LMs are dense, and all operations can be executed on specialized accelerators efficiently.
We expect that with advances in accelerated hardware, byte-level text processing will become a popular choice.

Of all the baseline models we reference, only \citet{transformerxl} uses recurrence to model arbitrary length history.
This technique could be added to tokenizer-free models as well.
Indeed, we expect this approach to be particularly well-suited to byte and character models where text gets mapped onto longer token sequences, as \citet{transformerxl} show that adding recurrence increases the length of context their model can effectively use.

\section{Extracting Word Representations}
\label{sec:representations}

\begin{figure*}[!htbp]
  \centering
  \includegraphics[width=1\textwidth]{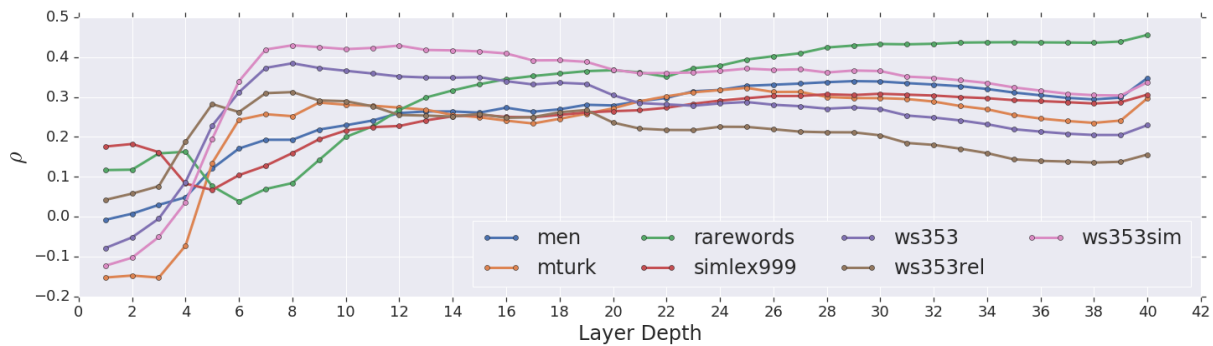}
  \caption{
  Performance on word similarity tasks described in \newcite{swivel}: Spearman's $\rho$ measuring correlation between human annotation and cosine similarities on word representations generated from activations on different transformer layers.
  }
  \label{fig:swivel}
\end{figure*}

In this section, we test our model's ability to produce meaningful word-level representations. We investigate this by feeding the model single words, and evaluating its intermediate activations on word similarity tasks.

Since our model is trained to predict each individual character, activations within a word only have partial information about that word.
To get a word representation, we append an empty space character at the end of the input word.
The activation at the space position from the transformer's feed-forward layer takes all characters into account, given the causal attention.
To predict what follows the space, the model must have a good understanding of the preceding word, so this activation can be used as a proxy for a word representation.

To evaluate our extracted word representations, we use the word similarity tasks described in Swivel~\cite{swivel}.
Following their evaluation methodology, we score word pairs using cosine similarity, and then measure the correlation with human ratings using Spearman's $\rho$.

We do not expect these results to be competitive, given that our model is never trained to represent words.
Moreover, the Swivel model is trained on a combination of Wikipedia and the Gigaword5 corpus~\cite{gigaword5} which is composed of 3.3 billion lowercased words with discarded punctuation.
They discard out-of-vocabulary words for evaluation, while we use all word pairs in the benchmark.
Nevertheless, this evaluation is valuable for comparing the \emph{relative} quality of representation across different layers.

Figure~\ref{fig:swivel} shows Spearman's $\rho$ across different layers of the model.
We observe two main phases of performance.
In the first phrase (layers 1-10), all task metrics improve with depth.
In the second phase (layers 11-40), performance either plateaus or degrades slightly with depth.
We suspect that the earlier layers learn general-purpose features which are linguistically relevant, while the final layers fine-tune specifically to the task of next character prediction.
Interestingly, the Rare Word and SimLex999 datasets do not follow this paradigm.
Their performance drops between layers 4-6, but picks up again and improves with depth (layers 6-40).
We hypothesize that the model may be storing words at different depths according to their frequency.
It would be interesting to investigate to what degree the improved performance of deeper LMs is due to better modeling of rare words/phrases.

Table~\ref{max_layer} shows the best performance of our model across all layers compared to the state-of-the-art model on word similarity.
The gap here is a reminder that work remains to be done on improving methods for extracting word representations from character models.

\begin{table}[h]
\begin{center}
\resizebox{\columnwidth}{!}{
\begin{tabular}{ c|cc } 
 Dataset & Ours & Swivel\\
 \hline
 men~\cite{bruni} & 0.35 &  0.76\\
 mturk~\cite{radinsky} & 0.32 & 0.72\\ 
 rarewords~\cite{luong} & 0.46 & 0.48\\ 
 simlex999~\cite{hill} & 0.31 & 0.40\\
 ws353~\cite{finkelstein} & 0.38 & - \\
 ws353rel~\cite{zesch} & 0.31 & 0.62\\
 ws353sim~\cite{agirre}& 0.43 & 0.75\\
\end{tabular}
}
\end{center}
\caption{Spearman's $\rho$ for different datasets using our model activations.
We report the best value achieved across all layers versus Swivel~\cite{swivel}.}
\label{max_layer}
\end{table}

\section{Conclusion}

We show that a tokenizer-free language model with sufficient capacity can achieve results that are competitive with word-based LMs.
Our model reads raw byte-level input without the use of any text preprocessing.
As such, the model has no direct access to word boundary information.
Finally, we show that our model's intermediate representations capture word-level semantic similarity and relatedness across layers.

\bibliography{emnlp-ijcnlp-2019}
\bibliographystyle{acl_natbib}

\end{document}